\documentclass[10pt,twocolumn,letterpaper]{article}

\usepackage{cvpr}
\usepackage{times}
\usepackage{epsfig}
\usepackage{graphicx}
\usepackage{amsmath}
\usepackage{amssymb}
\usepackage{float}

\usepackage{bm}
\usepackage{multirow,bigdelim}
\usepackage{pbox}
\usepackage{textcomp}
\usepackage{cite}
\usepackage{tabularx}
\usepackage{comment}
\usepackage{mathtools}
\usepackage{subcaption} 
\usepackage{enumitem}
\usepackage{algorithm}
\usepackage[noend]{algpseudocode}

\usepackage{pgfplots}
\pgfplotsset{compat=1.14}
\usetikzlibrary{shapes.geometric, arrows}

\definecolor{green}{RGB}{0,255,0}
\definecolor{orange}{RGB}{242, 150, 73}
\definecolor{fgreen}{RGB}{127,255,0}
\definecolor{cyan}{RGB}{127,255,255}
\definecolor{purple}{RGB}{230, 135, 255}
\definecolor{darkpurple}{RGB}{184,108,204}
\definecolor{yellow}{RGB}{251, 237, 65}
\definecolor{red}{RGB}{255, 40, 40}
\pgfplotscreateplotcyclelist{cycle-bar}{
	{black, thick, fill=cyan}, 
    {black, thick,fill=purple}, 
    {black, thick, fill=yellow},
    {black, thick, fill=red}
}
\pgfplotscreateplotcyclelist{cycle-graph}{
	{cyan!80!black, line width=1.25pt, mark=x, mark size=3, mark options={solid, cyan!50!black}}, 
    {purple!80!black, line width=1.25pt, mark=triangle*, mark options={solid, purple!50!black}}, 
    {yellow!80!black, line width=1.25pt, mark=*, mark options={yellow!50!black}},
    {red!80!black, line width=1.25pt, mark=square*, mark size=1.5, mark options={red!50!black}},
    {fgreen!80!black, line width=1.25pt, mark=diamond*, mark options={fgreen!50!black}},
    {orange!80!black, line width=1.25pt, mark=pentagon*, mark options={orange!50!black}}
}

\usepackage[breaklinks=true,bookmarks=false]{hyperref}

\cvprfinalcopy 

\def\cvprPaperID{794} 

\begin{document}

\title{Improving Object Localization with Fitness NMS and Bounded IoU Loss}

\author{Lachlan Tychsen-Smith, Lars Petersson\\
CSIRO (Data61)\\
CSIRO-Synergy Building, Acton, ACT, 2601\\
{\tt\small Lachlan.Tychsen-Smith@data61.csiro.au, \tt\small Lars.Petersson@data61.csiro.au}}

\maketitle

\begin{abstract}We demonstrate that many detection methods are designed to identify only a sufficently accurate bounding box, rather than the best available one. To address this issue we propose a simple and fast modification to the existing methods called Fitness NMS. This method is tested with the DeNet model and obtains a significantly improved MAP at greater localization accuracies without a loss in evaluation rate, and can be used in conjunction with Soft NMS for additional improvements. Next we derive a novel bounding box regression loss based on a set of IoU upper bounds that better matches the goal of IoU maximization while still providing good convergence properties. Following these novelties we investigate RoI clustering schemes for improving evaluation rates for the DeNet \textit{wide} model variants and provide an analysis of localization performance at various input image dimensions. We obtain a MAP of 33.6\%@79Hz and 41.8\%@5Hz for MSCOCO and a Titan X (Maxwell). Source code available from: \url{https://github.com/lachlants/denet}
\end{abstract}

\section{Introduction}
Multiclass object detection is defined as the joint task of localizing bounding boxes of instances and classifying their contents. In this paper, we address the problem of bounding box localization which has become increasingly important as the number of objects in view rises. In particular, by identifying a better fitting bounding box, current methods can better resolve the position, scale and number of unique instances in view. Modern object detection methods most often utilize a CNN\cite{lenet} based bounding box regression and classification stage followed by a Non-Max Suppression method to identify unique object instances. 

\begin{figure}[tb]
\centering
\begin{tikzpicture}
\node[inner sep=0pt, anchor=south west] (sample) at (0,0) {\includegraphics[width=.45\textwidth]{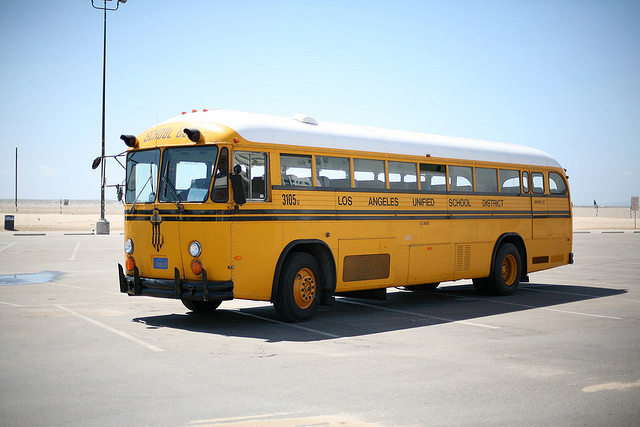}};
\begin{scope}[x={(sample.south east)},y={(sample.north west)}]
	\draw[red,ultra thick] (0.19,0.24) rectangle (0.9,0.75);
  	\node[red,anchor=west] (gt) at (0.15,0.80){\textbf{Groundtruth}};
	\draw[green!50!black,ultra thick] (0.45,0.28) rectangle (0.94,0.79);
  	\node[green!50!black,anchor=west] (bbox1) at (0.57,0.84){\textbf{IoU = 0.53}};
	\draw[blue,ultra thick] (0.16,0.26) rectangle (0.87,0.72);
 	\node[blue,anchor=west] (bbox2) at (0.43,0.19){\textbf{IoU = 0.83}};
\end{scope}
\end{tikzpicture}
\caption{MSCOCO\cite{mscoco} sample depicting the groundtruth bounding box (red) and two predicted bounding boxes (green and blue). Both predicted bounding boxes have a sufficient overlap with the groundtruth, i.e., $\mathrm{IoU} > 0.5$. Standard detection designs do not directly discriminate between the predicted boxes, whereas \textit{Fitness NMS} explicitly directs the model to select the blue box.}
\end{figure}
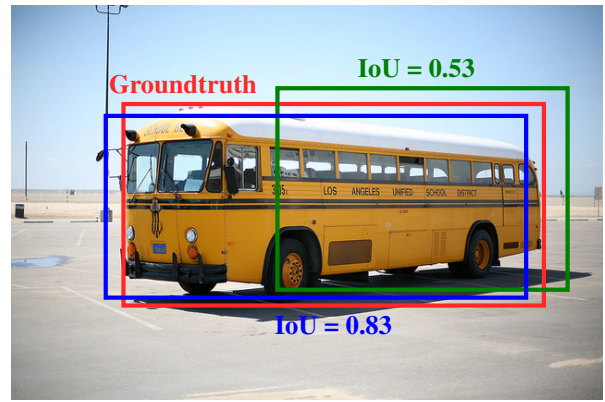

Introduced in R-CNN\cite{rcnn}, bounding box regression enables each region of interest (RoI) to estimate an updated bounding box with the goal of better matching the nearest true instance. Prior work has demonstrated that this task can be improved with multiple bounding box regression stages\cite{iterative-bbox}, increasing the number of (or carefully selecting) RoI anchors in the region proposal network\cite{ssd}\cite{retinanet}, and increasing the input image resolution (or using image pyramids)\cite{fpn}\cite{retinanet}. Alternatively, the DeNet\cite{denet} method demonstrated enhanced precision by improving the localization of the sampling RoIs before bounding box regression. This was achieved via a novel corner-based RoI estimation method, replacing the typical region proposal network (RPN)\cite{faster-rcnn} used in many other methods. This approach operates in real-time, demonstrated improved fine object localization and has the additional benefit of not requiring user-defined anchor bounding boxes. Despite only demonstrating our results with DeNet, we believe the novelties presented are equally applicable to other detectors. 


Despite its simplicity, Non-Max Suppression (see Algorithm \ref{algo:nms}) provides highly efficient detection clustering with the goal of identifying unique bounding box instances. Two functions $\mathrm{same}(.)$ and $\mathrm{score}(.)$, and the user defined constant $\bm{\lambda_{nms}}$ define the behaviour of this method. Typically,  $\mathrm{same}(b_i,b_j) = \mathrm{IoU}(b_i,b_j)$ and $\mathrm{score}(c,b_i) = \mathrm{Pr}(c|b_i)$ where $b_i,b_j$ are bounding boxes and $c$ is the corresponding class. Conceptually, $\mathrm{same}(.)$ uses the intersection-over-union to test whether two bounding boxes are best associated with the same true instance. While the $\mathrm{score}(.)$ function is used to select which bounding box should be kept, in this case, the one with the greatest confidence in the current class. Recent work has proposed decaying the associated detection \textit{score} instead of discarding detections\cite{soft-nms} and learning the NMS function directly from data\cite{learn-nms}\cite{learn-nms2}. 

In contrast, we investigate modifications to the $\mathrm{score}(.)$ and $\mathrm{same}(.)$ functions without changing the overall NMS algorithm. With \textit{Fitness NMS} we modify the $\mathrm{score}(.)$ function to better select bounding boxes which maximise their estimated IoU with the groundtruth, and with \textit{Bounded IoU Loss} we modify the training of the inputs to $\mathrm{same}(.)$ to provide tighter clusters. Following these novelties we provide an analysis of RoI clustering and input image sizes. 

\begin{algorithm}[tb]
\caption{Non-Max Suppression}\label{algo:nms}
\begin{algorithmic}[1]
\Procedure{NMS($B$,$c$)}{}
\State $B_{nms} \gets \emptyset$
\For {$b_i \in B$}
\State $discard \gets \mathrm{False}$
\For {$b_j \in B$}
\If {$\mathrm{same}(b_i, b_j) > \bm{\mathrm{\lambda_{nms}}}$}
\If {$\mathrm{score}(c,b_j) > \mathrm{score}(c,b_i)$} 
\State $discard \gets \mathrm{True}$
\EndIf
\EndIf
\EndFor
\If {\textbf{not} $discard$} 
\State $B_{nms} \gets B_{nms} \cup b_i$
\EndIf
\EndFor
\State \textbf{return} $B_{nms}$
\EndProcedure
\end{algorithmic}
\end{algorithm}

\section{Experimental Setup}

\subsection{Operating Parameters and Definitions}
Here we introduce some important parameters and definitions used throughout the paper:
\begin{itemize}
\item \textbf{Sampling Region-of-Interest (RoI)}: a bounding box generated before classification. Only applicable to two-stage detectors (e.g., R-CNN variants, etc).
\item \textbf{Intersection-over-Union (IoU)}: the intersection area divided by the union area for two bounding boxes. 
\item \textbf{Matching IoU} $\bm{\mathrm{\left ( \Omega \right )}}$: indicates the necessary IoU overlap between a model generated bounding box and a groundtruth bounding box before they are considered the same detection. A higher matching IoU indicates increased requirements for fine object localization. Typically, there are two matching IoUs, one used during training ($\bm{\mathrm{\Omega_{train}}}$) and one used for testing $\bm{\mathrm{\left ( \Omega_{test} \right )}}$. The training matching IoU ($\bm{\mathrm{\Omega_{train}}}$) is used to construct the target probability distribution for sampling RoIs, e.g., in DeNet given a ground-truth instance $(c_t,b_t)$ if $\mathrm{IoU}(b_s, b_t) \geq \bm{\mathrm{\Omega_{train}}}$ then $\mathrm{Pr}(c_t|b_s) = 1$. For Pascal VOC\cite{pascal-voc} $\bm{\mathrm{\Omega_{test}}}=0.5$ while MSCOCO\cite{mscoco} considers a range $\bm{\mathrm{\Omega_{test}}}\in [0.5,0.95]$. 
\item \textbf{Clustering IoU} $\bm{\mathrm{\left (\lambda_{nms} \right )}}$: indicates the IoU required between two detections before one is removed via Non Max Suppression, e.g., if $\mathrm{IoU}(b_0, b_1) > \bm{\mathrm{\lambda_{nms}}}$ and $\mathrm{Pr}(c|b_0) > \mathrm{Pr}(c|b_1)$ then the detector \textit{hit} $(c,b_1)$ is removed. In the DeNet model, the default value is $\bm{\mathrm{\lambda_{nms}}} = 0.5$, however, we vary this factor in some figures as indicated.
\end{itemize}

\subsection{Datasets, Training and Testing}

For validation experiments, we combine Pascal VOC\cite{pascal-voc} 2007 \texttt{trainval} and VOC 2012 \texttt{trainval} datasets to form the training data and test on Pascal VOC 2007 \texttt{test}. For testing, we train on MSCOCO\cite{mscoco} \texttt{trainval} and use the \texttt{test-dev} dataset for evaluation. Following DeNet\cite{denet} and SSD\cite{ssd}, all timing results are provided for an Nvidia Titan X (Maxwell) GPU with cuDNN v5.1 and a batch size of 8. Furthermore, unless stated otherwise, the same learning schedule, hyper-parameters, augmentation methods, etc were used as in the DeNet paper\cite{denet}.

\section{Fitness Non-Max Suppression}

In this section, we highlight a flaw in the Non-Max Supression based instance estimation, namely, the reliance on a single matching IoU. Following this analysis, we propose and demonstrate a novel \textit{Fitness NMS} algorithm. 

\subsection{Matching IoUs}

In Table \ref{table:match_ap} and \ref{table:match_ap_mscoco} we provide the MAP for the DeNet-34 (skip)\cite{denet} model trained with a variety of matching IoUs (keeping all other variables constant). We observe that increasing the train matching IoU can be used to significantly improve MAP at greater test matching IoUs at the expense of lower matching IoUs. In particular, we note an improvement of $1.2\%$ MAP@[0.5:0.95] for MSCOCO by selecting a training matching IoU near the center of the test matching IoU range. These results demonstrate a problem in many detection methods, i.e., they only operate optimally at a single matching IoU, when $\bm{\mathrm{\Omega_{train}}} = \bm{\mathrm{\Omega_{test}}}$. This property poses a problem when we want the detector to provide the \textit{best} available bounding box rather than just a \textit{sufficient} one, i.e., we wish to select the bounding box which maximizes IoU rather than just satisfying $\mathrm{IoU} > \bm{\mathrm{\Omega_{test}}}$. In the following, we develop the hypothesis that this problem stems from the applied instance estimation method, i.e., Non-Max Suppression acting on class probabilities. 

\begin{table}[tb]
\begin{center}
\begin{tabular}{c|c|c|c|c|c}
 & \multicolumn{5}{|c}{MAP@$\bm{\mathrm{\Omega_{test}}}$ (\%) }\\  
$\bm{\mathrm{\Omega_{train}}}$ & 0.5 & 0.6 & 0.7 & 0.8 & 0.9 \\
\hline
0.5 & \textbf{75.0} & 68.9 & 58.2 & 39.5 & 12.0 \\
0.7 & 73.7 & \textbf{69.0} & \textbf{60.6} & \textbf{45.8} & 17.2 \\
0.9 & 62.9 & 59.0 & 52.8 & 42.8 & \textbf{26.4}
\end{tabular}
\end{center}
\caption{VOC 2007 \texttt{test} results for DeNet-34 model with a variable matching IoU. The model only performs optimally when both training and testing IoUs are equal.}
\label{table:match_ap}
\end{table}

\begin{table}[tb]
\begin{center}
\begin{tabular}{c|ccc|ccc}
 & \multicolumn{3}{|c}{MAP@$\bm{\mathrm{\Omega_{test}}}$ (\%) } & \multicolumn{3}{|c}{MAP@Area (\%) }\\  
 $\bm{\mathrm{\Omega_{train}}}$ & 0.5:0.95 & 0.5 & 0.75 & S & M & L \\
\hline
0.5 & 29.5 & \textbf{47.9} & 31.1 & \textbf{8.8} & 30.9 & 47.0  \\
0.7 & \textbf{30.7} & 45.9 & \textbf{33.5}& 7.9 & \textbf{33.6} & \textbf{50.9} \\
\end{tabular}
\end{center}
\caption{Comparing MSCOCO \texttt{test-dev} results for DeNet-34 model with a training matching IoU of 0.5 and 0.7. Training with an IoU near the center of the testing IoU range improves MAP[0.5:0.95] by +1.2\%.}
\label{table:match_ap_mscoco}
\end{table}

\subsection{Detection Clustering}

To indicate the \textit{score} of a bounding box $b_j$, many detection models (including DeNet\cite{denet}) apply: 
\begin{gather}
\mathrm{score}(c,b_j) = \mathrm{Pr}(c | b_j)  
\label{eq:old_fitness}
\end{gather}
where $\mathrm{Pr}(c|b_j)$ provides the probability that instance $(c,b_j)$ is within the groundtruth assignment for the image. In practice, the distribution $\mathrm{Pr}(c | b_j)$ is trained to indicate whether a groundtruth instance $(c,b_t)$ has an overlap satisfying $\mathrm{IoU}(b_j, b_t) > \bm{\mathrm{\Omega_{train}}}$. Following class estimation, the Non-Max Suppression (NMS) algorithm is applied to $\mathrm{score}(c,b)$ over the bounding boxes $b$ to identify the most likely set of unique instance candidates (see Algorithm \ref{algo:nms}). In a typical production scenario, the clustering IoU $\bm{\mathrm{\lambda_{nms}}}$ is manually optimized to obtain the maximum recall within a predefined number of output detections over a validation set. In this context, the goal of the NMS algorithm is to simultaneously reduce the number of output detections (via deduplication) while maximizing the recall at the desired matching IoU (singular or range).

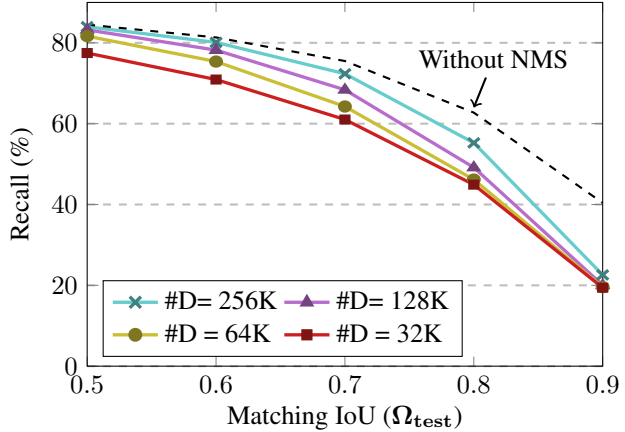
\begin{figure}[tb] 
\centering
\begin{tikzpicture}
	\begin{axis}[
    	yscale=0.85,
	    title style={yshift=3ex,align=center},
        grid style={thick, dashed},
		xlabel=Matching IoU ({$\bm{\mathrm{\Omega_{test}}}$}), xtick={0.5,0.6,0.7,0.8,0.9},
		xmin=0.5, xmax=0.9,
		ymajorgrids=true,
		ylabel=Recall (\%), ymin=0, ymax=90,
        cycle list name=cycle-graph,
        legend pos=south west,
		legend cell align=left,
        legend columns=2
	]

 	\addplot coordinates {(0.5,83.98) (0.6,80.11) (0.7,72.35) (0.8,55.25) (0.9,22.58)};
 	\addplot coordinates {(0.5,83.21) (0.6,78.20) (0.7,68.38) (0.8,49.16) (0.9,20.04)};
 	\addplot coordinates {(0.5,81.66) (0.6,75.37) (0.7,64.25) (0.8,46.17) (0.9,19.56)};
 	\addplot coordinates {(0.5,77.48) (0.6,70.92) (0.7,61.03) (0.8,44.91) (0.9,19.41)};
	\addplot[black, thick, dashed] coordinates {(0.5,84.51) (0.6,81.34) (0.7,75.50) (0.8,62.73) (0.9,40.40)};
	\legend{\#D= 256K, \#D= 128K, \#D = 64K, \#D = 32K}
  	\node[anchor=west] (upper) at (axis cs:0.75,76){Without NMS};
    \draw[->, thick](upper)--(axis cs:0.8,64);
	\end{axis}
\end{tikzpicture}
\caption{VOC 2007 \texttt{test} recall for DeNet-34 model with $\bm{\mathrm{\Omega_{train}}}=0.5$. The clustering IoU $\left (\bm{\mathrm{\lambda_{nms}}} \right)$ was varied to obtain the desired number of output detections (\textbf{\#D}). Note the large delta between the \textit{Without NMS} and other lines at high matching IoUs (e.g., $\bm{\mathrm{\Omega_{test}}}=0.9$).} 
\label{fig:cover_vs_nms}
\end{figure}

In Figure \ref{fig:cover_vs_nms}, we demonstrate the performance of the standard NMS algorithm with the DeNet-34 (skip) model at varying clustering IoUs ($\bm{\mathrm{\lambda_{nms}}}$) and matching IoUs ($\bm{\mathrm{\Omega_{test}}}$). Note that the \textit{Without NMS} line indicates an upper bound on recall and has an order of magnitude more detections, i.e., approx. 2.5M. We observe that this method performs well when $\bm{\mathrm{\Omega_{train}}}=\bm{\mathrm{\Omega_{test}}}=0.5$, quickly converging to the \textit{Without NMS} line as detections are added, e.g., there is a delta of approximately 7\% recall between the \textit{Without NMS} line and 32K detections. However, with greater localization accuracies $\bm{\mathrm{\Omega_{test}}} \gg 0.5$ the performance worsens and converges slowly as detections are added. In particular, we note the recall with $\bm{\mathrm{\Omega_{test}}}=0.9$ dropped from near 40\% to less than $20\%$ as detections were culled via the NMS algorithm. The large delta between the \textit{Without NMS} line and other results at high matching IoUs indicates that the correct bounding boxes are being identified during RoI prediction, however, the NMS algorithm is quickly discarding them. 

These results suggest that naively applying NMS with current detection models is suboptimal for high localization accuracies. We propose that this phenomenon \textit{partially} stems from the fact that the DeNet model is trained with a single matching IoU, i.e., during training $\mathrm{Pr}(c=c_t|b_s)=1$ if the sampling bounding box $b_s$ overlaps a groundtruth instance $(c_t,b_t)$ with $\mathrm{IoU}(b_s,b_t) \geq 0.5$. This makes the class distribution a poor discriminator between bounding boxes of better or worse IoU since it is trained to equal one whether the IoU is $0.5$ or $0.9$, etc. 


\subsection{Novel Fitness NMS Method}

To address the issues described in the previous sections we propose augmenting Equation \ref{eq:old_fitness} with an additional expected \textit{fitness} term:
\begin{align}
\label{eq:fitness}
\mathrm{score}(c,b_j) &=  \mathrm{Pr}(c | b_j) \mathrm{E} \left [ f_j | c \right] 
\end{align}
where $f_j$, referred to as \textit{fitness} in this paper, is a discrete random variable indicating the IoU overlap between the bounding box $b_j$ and any overlapping groundtruth instance, and $\mathrm{E}[f_j|c]$ is the expected value of $f_j$ given the class $c$. With this formulation, bounding boxes with both greater estimated IoU overlap and class probability will be favored in the clustering algorithm. 

In our implementation, the \textit{fitness} $f_j$ can take on $F$ values ($F=5$ in this paper) and is mapped via:
\begin{gather}
\rho_j = \max_{b \in B_T} \mathrm{IoU}(b_j, b) \\
f_j = \left \lfloor F (2 \rho_j - 1) \right \rfloor
\end{gather}
where $0 \leq \rho_j \leq 1$ is the maximum IoU overlap between $b_j$ and the set of groundtruth bounding boxes $B_T$. If $\rho_j$ is less than 0.5, the bounding box $b_j$ is assigned the \textit{null} class with no associated fitness. From these definitions the expected value is given by:
\begin{gather}
\mathrm{E} \left [ f_j | c \right] = \sum^{n < F}_{n=0} \lambda_n \mathrm{Pr}(f_j=n|c)
\end{gather}
where $\lambda_n = \frac{1}{2} + \frac{n}{2F}$ is the minimum IoU overlap associated with a fitness of $f_j=n$. In estimating $\mathrm{Pr}(f_j|c)$ from supervised training data we investigated two variants:
\begin{enumerate}[label=(\alph*)]
\item \textbf{Independent Fitness} $\left(\bm{\mathrm{I_F}} \right)$: Model $\mathrm{Pr}(c|b_j)$ and $\mathrm{Pr}(f_j|c)$ independently and assume fitness is independent of class, i.e., $\mathrm{Pr}(f_j|c) = \mathrm{Pr}(f_j)$. For training, we replaced the final layer of a pretrained DeNet model with one which estimates $\mathrm{Pr}(f_j|b_j)$ and $\mathrm{Pr}(c|b_j)$, both of which are jointly optimized via separate cross entropy losses. The method required one additional hyperparameter, the cost weighting for the fitness probability which was set to 0.1. With this method, the model produces $|C|+|F|+2$ outputs per RoI. 
\item \textbf{Joint Fitness} $\left(\bm{\mathrm{J_F}} \right)$: Estimate fitness and class jointly with $\mathrm{Pr}(c,f_j|b_j)$, and apply the following identity:
\begin{gather}
\mathrm{score}(c, b_j) = \sum^{n<F}_{n=0} \lambda_n \mathrm{Pr}(c, f_j=n | b_j)
\end{gather}
For training, we replaced the final layer of a pretrained DeNet model with one which jointly estimates $\mathrm{Pr}(c,f_j|b_j)$ instead of $\mathrm{Pr}(c|b_j)$. Note the softmax is normalised over both class and fitness. The training method required no additional hyperparameters and the same cost weighting was used as in the original layer. Since the \textit{null} class has no associated fitness, this method requires $|C||F| + 1$ model outputs per RoI. 
\end{enumerate}

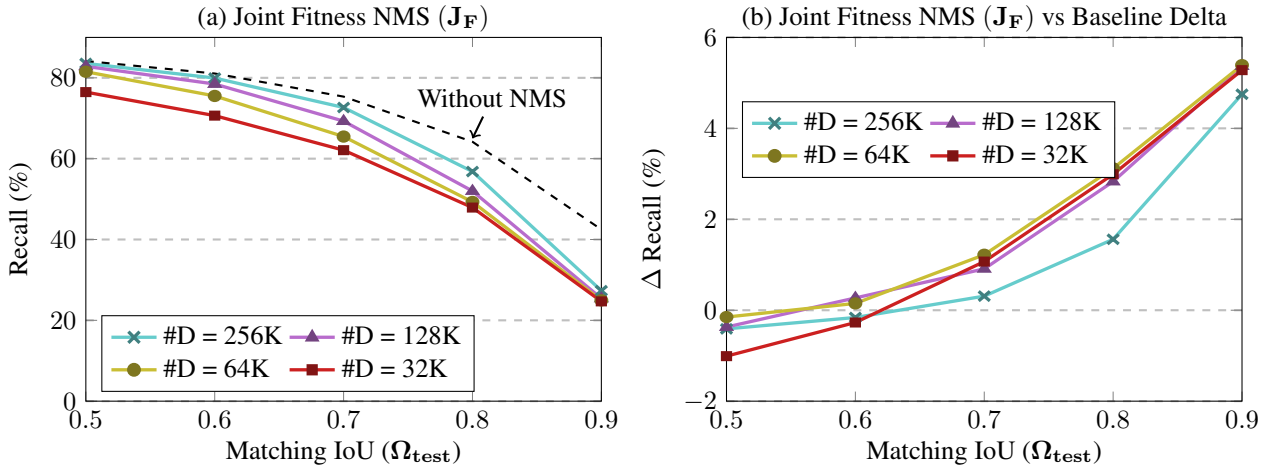
\begin{figure*}[tb] 
\centering
\scalebox{1}[1]{
\begin{tikzpicture}
	\begin{axis}[
		yscale=0.85,
	    title style={yshift=3ex,},
    	title= (a) Joint Fitness NMS $\left (\bm{\mathrm{J_F}} \right )$,
        grid style={thick, dashed},
		xlabel=Matching IoU ({$\bm{\mathrm{\Omega_{test}}}$}), xtick={0.5,0.6,0.7,0.8,0.9},
		xmin=0.5, xmax=0.9,
		ymajorgrids=true,
		ylabel=Recall (\%), ymin=0, ymax=90,
        cycle list name=cycle-graph,
        legend pos=south west,
		legend cell align=left,
        legend columns=2,
	]
	\addplot coordinates {(0.5,83.57) (0.6,79.95) (0.7,72.66) (0.8,56.81) (0.9,27.33)};
	\addplot coordinates {(0.5,82.84) (0.6,78.47) (0.7,69.29) (0.8,51.99) (0.9,25.41)};
	\addplot coordinates {(0.5,81.51) (0.6,75.52) (0.7,65.47) (0.8,49.29) (0.9,24.95)};
	\addplot coordinates {(0.5,76.47) (0.6,70.65) (0.7,62.10) (0.8,47.90) (0.9,24.69)};
	\addplot[black, thick, dashed] coordinates {(0.5,84.16) (0.6,81.07) (0.7,75.35) (0.8,64.18) (0.9,42.47)};
	\legend{ \#D = 256K, \#D = 128K, \#D = 64K, \#D = 32K}
  	\node[anchor=west] (upper) at (axis cs:0.75,75){Without NMS};
    \draw[->, thick](upper)--(axis cs:0.8,65);
\end{axis}
\end{tikzpicture}
\begin{tikzpicture}
	\begin{axis}[
		yscale=0.85,
	    title style={yshift=3ex,align=center},
	   	title= {(b) Joint Fitness NMS $\left (\bm{\mathrm{J_F}} \right)$ vs Baseline Delta},
        grid style={thick, dashed},
		xlabel=Matching IoU ({$\bm{\mathrm{\Omega_{test}}}$}), xtick={0.5,0.6,0.7,0.8,0.9},
		xmin=0.5, xmax=0.9,
		ymajorgrids=true,
		ylabel=$\Delta$ Recall (\%), ymin=-2, ymax=6,
        cycle list name=cycle-graph,
        legend pos=north west,
		legend cell align=left,
        legend columns=2
	]
	\addplot coordinates {(0.5,-0.41) (0.6,-0.16) (0.7,0.31) (0.8,1.56) (0.9,4.75)};
 	\addplot coordinates {(0.5,-0.37) (0.6,0.27) (0.7,0.91) (0.8,2.83) (0.9,5.37)};
 	\addplot coordinates {(0.5,-0.15) (0.6,0.15) (0.7,1.22) (0.8,3.12) (0.9,5.39)};
  	\addplot coordinates {(0.5,-1.01) (0.6,-0.27) (0.7,1.07) (0.8,2.99) (0.9,5.28)};
	\legend{\#D = 256K, \#D = 128K, \#D = 64K, \#D = 32K}
	\end{axis}
\end{tikzpicture}}
\caption{VOC 2007 \texttt{test} recall with DeNet-34: (a) with Joint Fitness NMS, and (b) delta obtain by switching from Joint Fitness NMS to standard NMS. We varied the clustering IoU to obtain the desired number of detections (\#D). Improved recall with IoU $>$ 0.6 was observed with Joint Fitness NMS enabled, peaking at approx. +5\% as $\bm{\mathrm{\Omega_{test}}}=0.9$.} 
\label{fig:fitness_nms_compare}
\end{figure*}

In Table \ref{table:voc_fitness}, we compare the MAP of a DeNet-34 (skip) model on Pascal VOC 2007 \texttt{test} augmented with the novel Fitness NMS methods over a range of matching IoUs ($\bm{\mathrm{\Omega_{test}}}$). The Baseline model was retrained starting from an already trained model such that all variants experienced the same number of training epochs.

The Joint Fitness NMS method demonstrates a clear lead at high matching IoUs with no observable loss for low matching IoUs. In Figure~\ref{fig:fitness_nms_compare}, we provide the recall for the Joint Fitness NMS method and the recall delta between Joint Fitness NMS and Baseline for the same model. These results directly demonstrate the improved recall at various operating points obtained by the Fitness NMS methods at fine localization accuracies with negligible losses for coarse localization. 

\begin{table}[tb]
\begin{center}
\begin{tabular}{l|c|c|c|c|c}
& \multicolumn{5}{|c}{MAP@$\bm{\mathrm{\Omega_{test}}}$ (\%) }\\  
NMS Method & 0.5 & 0.6 & 0.7 & 0.8 & 0.9 \\
\hline
Baseline & 75.3 & 70.4 & 61.0 & 45.3 & 18.4 \\
Ind. Fitness & 75.4 & 70.2 & 63.2 & 48.2 & 25.3 \\
Joint Fitness & \textbf{75.5} & \textbf{70.8} & \textbf{64.0} & \textbf{49.7} & \textbf{27.9}
\end{tabular}
\end{center}
\caption{ Pascal VOC 2007 \texttt{test} results for DeNet-34 (skip) with varying NMS methods and $\bm{\mathrm{\Omega_{test}}}$. Baseline is the original model and NMS method. Fitness NMS methods improves MAP at high IoU ($\bm{\mathrm{\Omega_{test}}}$).}
\label{table:voc_fitness}
\end{table}

\begin{table}[tb]
\begin{center}
\begin{tabular}{l|c|ccc}
 & Eval. & \multicolumn{3}{|c}{MAP@$\bm{\mathrm{\Omega_{test}}$} (\%)} \\
 Model & Rate &0.5:0.95 & 0.5 & 0.75\\
\hline 
DeNet-34 $\bm{\mathrm{S}}$ & 82 Hz & 29.5 & \textbf{47.9} & 31.1 \\
DeNet-34 $\bm{\mathrm{S}}$+$\bm{\mathrm{I_F}}$ & 81 Hz & 31.1 & 47.7 & 33.1 \\
DeNet-34 $\bm{\mathrm{S}}$+$\bm{\mathrm{J_F}}$ & 78 Hz & \textbf{31.4} & \textbf{47.9} & \textbf{33.5} \\
\hline
DeNet-34 $\bm{\mathrm{W}}$ & 44 Hz & 30.0 & 48.9 & 31.8 \\
DeNet-34 $\bm{\mathrm{W}}$+$\bm{\mathrm{I_F}}$ & 44 Hz & 32.1 & 48.5 & 34.2 \\
DeNet-34 $\bm{\mathrm{W}}$+$\bm{\mathrm{J_F}}$ & 44 Hz & \textbf{32.6} & \textbf{49.2} & \textbf{34.9} \\
\hline
DeNet-101 $\bm{\mathrm{W}}$ & 17 Hz & 33.8 & 53.4 & 36.1 \\
DeNet-101 $\bm{\mathrm{W}}$+$\bm{\mathrm{I_F}}$ & 17 Hz & 35.2 & 52.9 & 37.7 \\
DeNet-101 $\bm{\mathrm{W}}$+$\bm{\mathrm{J_F}}$ & 17 Hz & \textbf{36.5 }& \textbf{53.9} & \textbf{39.1} \\
\end{tabular}
\end{center}
\caption{MSCOCO \texttt{test-dev} results with varying NMS methods. Models are designated $\bm{\mathrm{S}}$ for the \textit{skip} variant and $\bm{\mathrm{W}}$ for the \textit{wide} variant, and $\bm{\mathrm{I_F}}$ utilize the \textit{independent} variant and $\bm{\mathrm{J_F}}$ the \textit{joint} variant of Fitness NMS. Our NMS methods improve MAP when $\bm{\mathrm{\Omega_{test}}}=0.75$.}
\label{table:mscoco_fitness}
\end{table}

In Table \ref{table:mscoco_fitness}, we provide MSCOCO \texttt{test-dev} results for DeNet-34 (skip), DeNet-34 (wide) and DeNet-101 (wide) models augmented with the Fitness NMS methods. The Independent Fitness method consistently demonstrated a marginal loss in precision at $\bm{\mathrm{\Omega_{test}}}=0.5$ with a significant improvement at $\bm{\mathrm{\Omega_{test}}}=0.75$, however, the Joint Fitness NMS method demonstrated no measurable loss at $\bm{\mathrm{\Omega_{test}}}=0.5$ and provided a greater improvement for $\bm{\mathrm{\Omega_{test}}}=0.75$. In particular, these methods provide a significant improvement for large objects which, most likely, have many associated sampling RoIs. The DeNet-101 model produced a large improvement moving from the \textit{independent} to \textit{joint} variant, suggesting that the increased modelling \textit{capacity} it provides is better utilized with the \textit{joint} method. Overall, both Fitness NMS methods produced a significant improvement for MAP@[0.5:0.95] between 1.9-2.7\%. 

\newpage

A minor loss in evaluation rate was observed for Joint Fitness NMS due to the increased number of values in the final softmax (401 vs 81). Applying the \textit{joint} variant to a single-stage detector\cite{ssd}\cite{retinanet} would likely result in a greater evaluation rate penalty due to much greater number of softmax's required in current designs (one for each candidate RoI). In these applications, the \textit{independent} variant might be a better candidate.
\subsubsection{Applying Soft-NMS}
In Table \ref{table:mscoco_fitness_softnms} we demonstrate that the Fitness NMS methods can be used in parallel with the Soft NMS method\cite{soft-nms}. Our results indicate an improvement between 0.7 and 1.3\% MAP for models utilizing both the standard and Fitness NMS methods. This highlights that the Soft NMS and Fitness NMS methods address tangential problems. Note that subsequent models in the paper do \textbf{not} use Soft NMS.

\begin{table}[tb]
\begin{center}
\begin{tabular}{l|c|ccc}
 & Eval. & \multicolumn{3}{|c}{MAP@$\bm{\mathrm{\Omega_{test}}$} (\%)} \\
 Model & Rate &0.5:0.95 & 0.5 & 0.75\\
\hline 
DeNet-34 $\bm{\mathrm{S}}$ & 80 Hz & 30.4 & \textbf{48.4} & 32.6 \\
DeNet-34 $\bm{\mathrm{S}}$+$\bm{\mathrm{J_F}}$ & 81 Hz & \textbf{32.2} & 48.1 & \textbf{34.8} \\
\hline
DeNet-34 $\bm{\mathrm{W}}$ & 43 Hz & 31.0 & 49.4 & 33.2 \\
DeNet-34 $\bm{\mathrm{W}}$+$\bm{\mathrm{J_F}}$ & 41 Hz & \textbf{33.4} & \textbf{49.7} & \textbf{36.0} \\
\hline
DeNet-101 $\bm{\mathrm{W}}$ & 17 Hz & 35.1 & 50.9 & 37.8 \\
DeNet-101 $\bm{\mathrm{W}}$+$\bm{\mathrm{J_F}}$ & 15 Hz & \textbf{37.2} & \textbf{54.2} & \textbf{40.2} 
\end{tabular}
\end{center}
\caption{Applying Gaussian variant ($\sigma=0.5$) of Soft NMS\cite{soft-nms} method to models in Table \ref{table:mscoco_fitness}. MAP improved by 0.7-1.3\% for both standard and Joint Fitness NMS models.}
\label{table:mscoco_fitness_softnms}
\end{table}

\section{Bounding Box Regression}

In this section, we discuss the role of bounding box regression in detection models. We then derive a novel bounding box regression loss based on a set of upper bounds which better matches the goal of maximising IoU while maintaining good convergence properties (smoothness, robustness, etc) for gradient descent optimizers. 

\subsection{Same-Instance Classification}

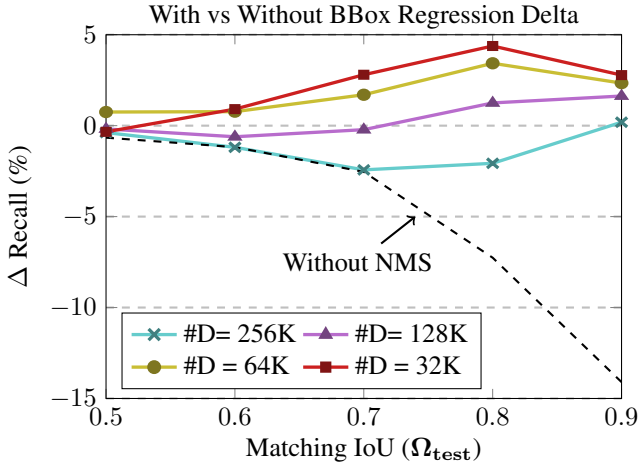
\begin{figure}[tb] 
 \centering
\begin{tikzpicture}
	\begin{axis}[
    	yscale=0.85,
	    title style={yshift=3ex,align=center},
     	title= {With vs Without BBox Regression Delta},
        grid style={thick, dashed},
		xlabel=Matching IoU ({$\bm{\mathrm{\Omega_{test}}}$}), xtick={0.5,0.6,0.7,0.8,0.9},
		xmin=0.5, xmax=0.9,
		ymajorgrids=true,
		ylabel=$\Delta$ Recall (\%), ymin=-15, ymax=5,
        cycle list name=cycle-graph,
        legend pos=south west,
		legend cell align=left,
        legend columns=2
	]

	\addplot coordinates {(0.5,-0.38) (0.6,-1.19) (0.7,-2.43) (0.8,-2.07) (0.9,0.19)};
	\addplot coordinates {(0.5,-0.18) (0.6,-0.61) (0.7,-0.22) (0.8,1.25) (0.9,1.63)};
	\addplot coordinates {(0.5,0.75) (0.6,0.77) (0.7,1.7) (0.8,3.43) (0.9,2.34)};
	\addplot coordinates {(0.5,-0.34) (0.6,0.91) (0.7,2.8) (0.8,4.38) (0.9,2.78)};
	\addplot[black, thick, dashed] coordinates {(0.5,-0.66) (0.6,-1.18) (0.7,-2.54) (0.8,-7.26) (0.9,-14.1)};
	\legend{\#D= 256K, \#D= 128K, \#D = 64K, \#D = 32K}
  	\node[anchor=west] (upper) at (axis cs:0.63,-7.5){Without NMS};
    \draw[->, thick](upper)--(axis cs:0.74,-5);
	\end{axis}
\end{tikzpicture}
\caption{VOC 2007 \texttt{test} recall delta for DeNet-34 (skip) with and without bounding box regression. We varied the clustering IoU $\left( \bm{\mathrm{\lambda_{nms}}}\right )$ to obtain the desired number of detections (\textbf{\#D}). Bounding box regression provides improved recall for high matching IoUs and low detection numbers.} 
\label{fig:cover_no_bbox}
\end{figure}

Since its introduction in R-CNN\cite{rcnn}, bounding box regression has provided improved localization for many two-stage\cite{faster-rcnn}\cite{fpn}\cite{deform-rfcn}\cite{denet} and single-stage detectors\cite{retinanet}\cite{ssd}\cite{yolo}. With bounding box regression each RoI is able to identify an updated bounding box which better matches the nearest ground-truth instance. Due to the improved accuracy of the sampling RoIs in DeNet compared to R-CNN derived methods, it is natural to assume a reduced need for bounding box regression. However, we demonstrate the contrary in Figure \ref{fig:cover_no_bbox}, which provides the change in recall obtained with the model in Figure \ref{fig:cover_vs_nms} with and without bounding box regression enabled. We observe improved recall with bounding box regression enabled for most data points except notably the \textit{Without NMS} and 256K lines at high matching IoUs. These results indicate that the RoIs have improved recall, however, as NMS is applied (other lines) the regressed bounding boxes perform better. Furthermore, these results demonstrate that bounding box regression is still important in the DeNet model under typical operating conditions (red and yellow lines). 

We propose that the improved recall obtained by the bounding box regression enabled DeNet model is \textbf{not} primarily due to improving the position and dimensions of individual bounding boxes, but instead due to its interaction with the NMS algorithm. That is, to test if two RoIs $(b_0,b_1)$ are best associated with the same instance, we check if $\mathrm{IoU}(\beta_0,\beta_1) > \bm{\mathrm{\lambda_{nms}}}$ where $(\beta_0,\beta_1)$ are the associated updated bounding boxes (after bounding box regression). Therefore, if $b_0$ and $b_1$ share the same nearest groundtruth instance they are significantly more likely to satisfy the comparison with bounding box regression enabled. As a result, bounding box regression is particularly important for estimating unique instances in cluttered scenes where sampling RoIs can easily overlap multiple true instances. 

\subsection{Bounded IoU Loss} 
Here we propose a novel bounding box loss and compare it to the R-CNN method used so far. This new loss aims to maximise the IoU overlap between the RoI and the associated groundtruth bounding box, while providing good convergence properties for gradient descent optimizers. 

Given a sampling RoI $b_s=(x_s,y_s,w_s,h_s)$, an associated groundtruth target $b_t=(x_t,y_t,w_t,h_t)$, and an estimated bounding box $\beta=(x,y,w,h)$ the widely used R-CNN formulation\cite{fast-rcnn} provides the following cost functions:
\begin{align}
\mathrm{Cost_x} &= L_1 \left (\frac{\Delta x }{w_s } \right ) \\
\mathrm{Cost_w} &= L_1 \left (\ln \left (\frac{w}{w_t} \right ) \right )
\end{align}
where $\Delta x = x-x_t$ and $L_1(z)$ is the Huber Loss\cite{huber-loss} (also known as smooth L1 loss\cite{fast-rcnn}).  Note that we restrict this analysis to the X position and width for the sake of brevity, the Y position and height equations can be identified with suitable substitutions. The Huber loss is defined by:
\begin{gather}
L_{\tau}(z) = \left \lbrace \begin{matrix} \frac{1}{2} z^2 & |z| < \tau\\
\tau |z|-\frac{1}{2} \tau^2 & \mathrm{otherwise}
\end{matrix} \right.
\end{gather}
Ideally, we would train the bounding box regression to directly minimize the IoU, e.g., $\mathrm{Cost}=L_1(1-\mathrm{IoU}(b,b_t))$, however, CNN's struggle to minimize this loss under gradient descent due to the high non-linearity, multiple degrees of freedom and existence of multiple regions of zero gradient of the IoU function. As an alternative, we consider the set of cost functions defined by:
\begin{gather}
\mathrm{Cost_i} = 2 L_1( 1-\mathrm{IoU_{B}}(i, b_t)) 
\label{eq:local_cost}
\end{gather}
where $\mathrm{IoU}(b,b_t) \leq \mathrm{IoU_{B}}(i, b_t)$ is an upper bound of the IoU function with the free parameter $i \in \{ x,y,w,h \}$. We construct the upper bound by considering the maximum IoU attainable with the unconstrained free parameter, e.g., $\mathrm{IoU_{B}}(x, b_t)$ provides the IoU as a function of $x$ when $y=y_t$, $w=w_t$ and $h=h_t$. We obtain the following bounds on the IoU function:
\begin{align}
\mathrm{IoU_{B}}(x, b_t) =& \max \left ( 0, \frac{w_t - 2 |\Delta x|} {w_t + 2 |\Delta x|} \right )
\label{eq:local_x}
\\
\mathrm{IoU_{B}}(w, b_t) =& \min \left (\frac{w} {w_t}, \frac{w_t}{w} \right) 
\label{eq:local_w}
\end{align}

\begin{figure*}[tb] 
\centering
\begin{tikzpicture}
	\begin{axis}[
   		yscale=0.85,
	    title style={yshift=3ex,align=center},
	    height=6.5cm,
        width=0.45\textwidth,
        grid style={thick, dashed},
		xlabel= {$\Delta x / w_t$}, 
		xmin=-0.5, xmax=0.5, xtick={-0.5,-0.25,0,0.25,0.5},
		ymajorgrids=true,
		ylabel= {$\mathrm{Cost_x}$}, 
        ymin=0, ymax=1,
        cycle list name=cycle-graph,
        no markers,
	]

	\addplot[smooth,very thick,cyan!80!black] {(1-(1-2*abs(x))/(1+2*abs(x)))^2};    
	\addplot[smooth,very thick,purple!80!black] {0.5*x*x};        
    \addplot[mark=none, red, dashed, very thick] coordinates {(0.1666666,0) (0.1666666,2)};
    \addplot[mark=none, red, dashed, very thick] coordinates {(-0.1666666,0) (-0.1666666,2)};
	\end{axis}
\end{tikzpicture}
\begin{tikzpicture}
	\begin{axis}[
    	yscale=0.85,
	    title style={yshift=3ex,align=center},
    	height=6.5cm,
        width=0.45\textwidth,
        grid style={thick, dashed},
		xlabel= {$w / w_t$}, 
		xmin=0, xmax=2.5,
		ymajorgrids=true,
		ylabel= {$\mathrm{Cost_w}$}, 
        ymin=0, ymax=1,
        cycle list name=cycle-graph,
        no markers,
	]

  	\addplot[smooth,very thick,cyan!80!black,domain=0.01:2.5] {(1 - min(x,1/x))^2};     	\addplot[smooth,very thick,purple!80!black,domain=0.01:2.5] {0.5*(ln(x))^2};       
    \addplot[mark=none, red, dashed, very thick] coordinates {(0.5,0) (0.5,1)};
    \addplot[mark=none, red, dashed, very thick] coordinates {(2,0) (2,1)};
	\end{axis}
\end{tikzpicture}
\caption{Comparing R-CNN bounding box regression loss (purple) and proposed Bounded IoU Loss (blue). Red dotted lines indicate the typical operating range given $\mathrm{IoU} \geq 0.5$. Bounded IoU Loss places a much greater emphasis on position.} 
\label{fig:cost_compare}
\end{figure*}
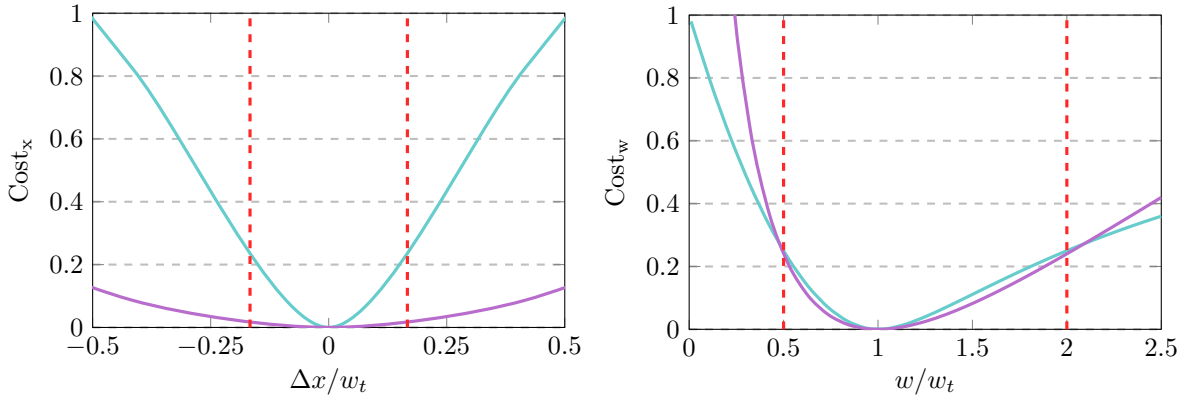

In Figure \ref{fig:cost_compare}, we compare the original R-CNN cost functions and the proposed Bounded IoU Loss. Since bounding box regression is only applied to boxes which have $\mathrm{IoU}(b_s, b_t) \geq 0.5$, the following constraints apply:
\begin{gather}
\frac{|x_s - x_t|}{w_t} \leq \frac{1}{6} \\
\frac{1}{2} \leq \frac{w_s}{w_t} \leq 2
\end{gather}

Note that these constraints can be obtained by applying the position and scale upper bounds derived above and the identity $\mathrm{IoU_B} \geq \mathrm{IoU} \geq 0.5 \Rightarrow \mathrm{IoU_B} \geq 0.5$. Assuming that the estimated bounding box $\beta$ has a better IoU than the sampling bounding box $b_s$, we provide these bounds in Figure \ref{fig:cost_compare}. We observe that the proposed cost for bounding box width has a nearly identical shape to the one used in the original algorithm, however, the proposed positional cost has a different shape and is considerably larger across the operating range. Furthermore, the positional cost is robust in the sense that as $|\Delta x| \rightarrow \infty$ the cost approaches a constant rather than diverging like the original cost, this may prove advantageous for applications which have large positional deltas. This analysis suggests that with the aim of maximizing IoU, a greater emphasis should be placed on position than the original R-CNN cost implies. Though not tested in this paper, we note that the R-CNN positional cost can be changed to $\mathrm{Cost_x} = 15 L_{0.16} \left (\Delta x / w_s  \right )$ to obtain a very similar shape over the defined operating range.

In Table \ref{table:bbox_reg}, we demonstrate the performance of the DeNet models with the Bounded IoU Loss ($\bm{\mathrm{B_{IoU}}}$). The model training was identical to the Joint Fitness models except with the bounding box regression cost factor changed from 1.0 to 0.125. This value was determined by matching the magnitude of the bounding box regression training loss after a few iterations to the value obtained previously (no parameter search was used). We observe a consistent improvement across varying matching thresholds ($\bm{\mathrm{\Omega_{test}}}$) for all models, particularly the DeNet-34 variants.

\begin{table}[tb]
\begin{center}
\begin{tabular}{l|ccc}
 & \multicolumn{3}{|c}{MAP@$\bm{\mathrm{\Omega_{test}}$} (\%)} \\
 Model & 0.5:0.95 & 0.5 & 0.75\\
\hline
DeNet-34 $\bm{\mathrm{S}}$+$\bm{\mathrm{J_F}}$ & 31.4 & 47.9 & 33.5 \\
DeNet-34 $\bm{\mathrm{S}}$+$\bm{\mathrm{J_F}}$+$\bm{\mathrm{B_{IoU}}}$ & \textbf{32.1} & \textbf{48.6} & \textbf{34.3} \\
\hline
DeNet-34 $\bm{\mathrm{W}}$+$\bm{\mathrm{J_F}}$ & 32.6 & 49.2 & 34.9 \\
DeNet-34 $\bm{\mathrm{W}}$+$\bm{\mathrm{J_F}}$+$\bm{\mathrm{B_{IoU}}}$ & \textbf{33.6} & \textbf{50.2} & \textbf{36.0} \\
\hline
DeNet-101 $\bm{\mathrm{W}}$+$\bm{\mathrm{J_F}}$ & 36.5 & 53.9 & 39.1 \\
DeNet-101 $\bm{\mathrm{W}}$+$\bm{\mathrm{J_F}}$+$\bm{\mathrm{B_{IoU}}}$ & \textbf{37.0} & \textbf{54.3} & \textbf{39.5} 
\end{tabular}
\end{center}
\caption{MSCOCO \texttt{test-dev} results with novel Bounded IoU Loss ($\bm{\mathrm{B_{IoU}}}$) for bounding box regression. Our novel loss consistently improved MAP across all categories.}
\label{table:bbox_reg}
\end{table}

\section{RoI Clustering} 

The \textit{wide} variant DeNet models generate a corner distribution with a $128\times 128$ spatial resolution producing 67M candidate bounding boxes\cite{denet}. To accommodate the increased numbers of candidates the original DeNet model increased the number of RoIs from $576$ to $2304$, resulting in a significant detrimental effect on the evaluation rate, e.g., DeNet-34 (wide) observed a near 50\% reduction in evaluation rate over the original model. To ameliorate this issue, we investigated two simple and fast clustering methods, one applying NMS on the corner distribution and the other applying NMS to the output bounding boxes. In the corner NMS method, when a corner at position $(x,y)$ of type $k$ is identified, i.e., $\mathrm{Pr}(s|k,y,x) > \lambda_C$, the algorithm checks if it has the maximum probability in a local $(2m+1)\times (2m+1)$ region.

\begin{table}[tb]
\begin{center}
\begin{tabular}{l|c|ccc}
 & Eval. & \multicolumn{3}{|c}{MAP@$\bm{\mathrm{\Omega_{test}}}$ (\%)}  \\
 Model & Rate &0.5:0.95 & 0.5 & 0.75  \\
\hline
DeNet-34 $\bm{\mathrm{W}}$ & 71 Hz & 33.1 & 48.7 & 35.6 \\
DeNet-34 $\bm{\mathrm{W}}$+$\bm{\mathrm{C}}$ & 79 Hz & 33.6 & 49.8 & \textbf{36.0} \\
DeNet-34 $\bm{\mathrm{W}}$+$\bm{\mathrm{N}}$ & 59 Hz & \textbf{33.7} & \textbf{50.7} & 35.8 \\
\hline
DeNet-101 $\bm{\mathrm{W}}$ & 22 Hz & 36.3 & 52.8 & 39.0 \\
DeNet-101 $\bm{\mathrm{W}}$+$\bm{\mathrm{C}}$ & 24 Hz & \textbf{36.9} & 53.9 & \textbf{39.5} \\
DeNet-101 $\bm{\mathrm{W}}$+$\bm{\mathrm{N}}$ & 21 Hz & \textbf{36.9} & \textbf{54.6} & 39.3 \\
\end{tabular}
\end{center}
\caption{MSCOCO \texttt{test-dev} results comparing clustering methods; $\bm{\mathrm{C}}$ indicates corner clustering with $m=1$, $\bm{\mathrm{N}}$ indicates standard NMS clustering with 0.7 threshold. Models were tested with 576 RoIs vs 2304 in other results. Both clustering methods improved MAP.}
\label{table:mscoco_corner_cluster}
\end{table}
\begin{table*}[tb]
\begin{center}
\begin{tabular}{l|c|c|ccc|ccc}
 & Mean Input & Eval. & \multicolumn{3}{|c|}{MAP@$\bm{\mathrm{\Omega_{test}}}$ (\%)} & \multicolumn{3}{|c}{MAP@Area (\%)} \\
Model & Pixels & Rate & 0.5:0.95 & 0.5 & 0.75 & S & M & L\\
\hline
FPN\cite{fpn} & 0.91 M & - & 36.2 & 59.1 & 39.0 & 18.2 & 39.0 & 48.2\\
D-RFCN\cite{r-fcn} & 0.51 M & - & 36.1& 56.7 & - & 14.8 & 39.8 & 52.2\\
Soft NMS\cite{soft-nms} & 0.91 M & - & 38.4 & 60.1 & - & 18.5 & 41.6 & 52.5\\
RetinaNet\cite{retinanet} & 0.91 M & 5 Hz & 39.1 & 59.1& 42.3 & 21.8 & 42.7 & 50.2\\
\hline
\textbf{Ours} ($\times 384$) & 0.11 M & 37 Hz & 32.3 & 47.9 & 34.5 & 8.7 & 33.7 & 54.5 \\
\textbf{Ours} ($\times 512$) & 0.19 M & 24 Hz & 36.9 & 53.9 & 39.5 & 13.8 & 39.3 & 56.5 \\
\textbf{Ours} ($\times 768$) & 0.42 M & 11 Hz & 39.5 & 58.0 & 42.6 & 18.9 & 43.5 & 54.1 \\
\textbf{Ours} ($\times 1024$) & 0.75 M & 6 Hz & 38.8 & 58.3 & 41.8 & 21.3 & 43.4 & 49.4 \\
\textbf{Ours} ($\times 1280$) & 1.18 M & 4 Hz & 36.3 & 56.2 & 39.0 & \textbf{22.1} & 42.1 & 42.3 \\
\textbf{Ours} ($\times 1536$) & 1.69 M & 3 Hz & 32.9 & 52.5 & 35.0 & 22.0 & 39.9 & 34.4 \\
\hline
\textbf{Ours} ($\times 512,\times 1024$) & 0.94 M & 5 Hz & \textbf{41.8} & \textbf{60.9} & \textbf{44.9} & 21.5 & \textbf{45.0} & \textbf{57.5} \\
\end{tabular}
\end{center}
\caption{MSCOCO \texttt{test-dev} results for DeNet-101 model with Joint Fitness NMS, Bounded IoU Loss and corner based RoI clustering. The maximum input dimension size was varied over 384, 512, 768, 1024, 1280 and 1536 with the ratio between the number of sampling RoIs and mean input pixels kept constant. For comparison at a similar pixel density and evaluation rate as RetinaNet we merged the $\times 512$ and $\times 1024$ results to obtain the bottom row.}
\label{table:mscoco_input_scaling}
\end{table*}



In Table \ref{table:mscoco_corner_cluster}, we compare the corner clustering method, standard NMS method (with a 0.7 threshold), and no clustering when the number of RoIs is reduced to 576 for the DeNet wide models. The results demonstrate a 30\% to 80\% improvement in evaluation rate due to the decreased number of RoI classifications needed. We found both clustering methods improved upon no clustering and obtained very similar MAP@[0.5:0.95] results, however, the standard NMS method was slightly slower due to an increased CPU load. Relative to the NMS method, the corner clustering method appears to perform slightly better at high matching IoU and worse at low matching IoUs.

\section{Input Image Scaling} \label{sec:input_image_scaling}

So far our model has been demonstrated at low input image size, i.e., rescaling the \textit{largest} input dimension to $512$ pixels. In comparison, state-of-the-art detectors typically use an input image with the \textit{smallest} side resized to 600 or 800 pixels. This smaller input image has provided our model with an improved evaluation rate at the cost of object localization accuracy. In the following, we relax the constraint on evaluation rate to demonstrate localization accuracies when computational resources are less constrained. 

In Table \ref{table:mscoco_input_scaling}, we provide MSCOCO results for the DeNet-101 (wide) model with Fitness NMS, Bounded IoU Loss and Corner Clustering with the largest input image dimension varied from 384 to 1536. Note that the model was not retrained with these settings, only tested. Since the benchmarked methods use different input image scaling methods, we provide the mean input pixels per sample (ignoring black borders) calculated over the MSCOCO \texttt{test-dev} dataset. These results demonstrate that input image size is very important for small and medium sized object localization in MSCOCO, however, we observed a loss in precision for large objects as scale is increased. This precision asymmetry suggests that multi-scale evaluation is likely optimal with the current design. 



\section{Conclusion}



\begin{figure}[tb] 
\centering
\begin{tikzpicture}
	\begin{axis}[
        grid style={thick, dashed},
		xlabel= {Inference time (ms)}, 
		xmin=0, xmax=250, 
		ymajorgrids=true,
 		ylabel= {MAP (\%)}, 
        ymin=20, ymax=45,
        cycle list name=cycle-graph,
        legend pos=south east,
		legend cell align=left,
        legend columns=1
	]

	\addplot coordinates {(209,41.8) (92.6,39.5) (42.4,36.9) (23.25,35.5)  (12.7,33.6)};
    \addlegendentry{\textbf{Ours}}

    \addplot coordinates {(58.8,33.8) (30.3,32.3) (12.2,29.5) };   
    \addlegendentry{DeNet\cite{denet}}

	\addplot coordinates {(198,39.1) (198,37.8) (154,37.1) (122,36.0) (90,34.4) (81,31.9)};
    \addlegendentry{RetinaNet\cite{retinanet}}

	\addplot coordinates {(152,33.2) (65,28.0)};
    \addlegendentry{DSSD\cite{dssd}}
\end{axis}
\end{tikzpicture}
\caption{MSCOCO \texttt{test-dev} MAP vs Inference Time comparison selecting top DeNet-34 and DeNet-101 results.}
\label{fig:mscoco_comp}
\end{figure}
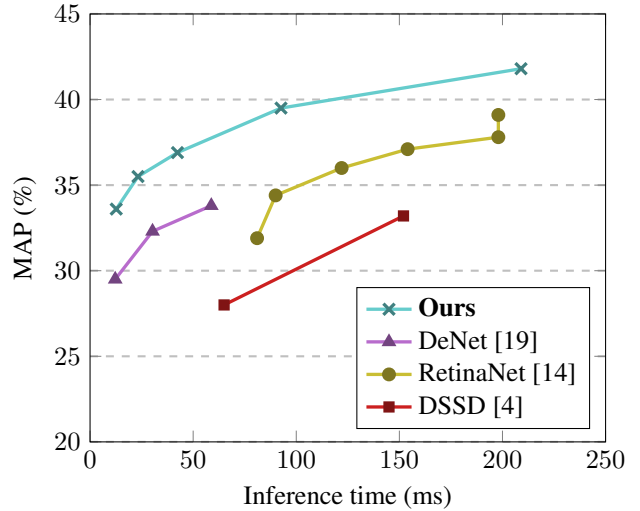

We highlighted an issue in common detector designs and propose a novel \textit{Fitness NMS} method to address it. This method significantly improves MAP at high localization accuracies without a loss in evaluation rate. Following this we derive a novel bounding box loss better suited to IoU maximisation while still providing convergence properties suitable to gradient descent. Combining these results with a simple RoI clustering method, we obtain highly competitive MAP vs Inference Times on  MSCOCO (see Figure \ref{fig:mscoco_comp}). These results highlight that with these modification a two-stage detector can be made highly competitive with single-stage methods in terms of MAP vs inference time. Though not demonstrated empirically in this paper, we believe the novelties presented in this paper are equally applicable to other bounding box detector designs. 

\newpage
{\small
\bibliographystyle{ieee}
\bibliography{egbib}
}

\end{document}